\documentclass[conference,letterpaper]{IEEEtran}

\IEEEoverridecommandlockouts                              

\usepackage{graphicx}
\usepackage{subfigure}
\usepackage{cite}
\usepackage{amsmath}
\usepackage{float}
\usepackage{threeparttable}
\usepackage{multirow}
\usepackage[left=0.71in,top=0.94in,right=0.71in,bottom=1.18in]{geometry}
\title{\huge
Large-scale Multi-modal Person Identification in Real Unconstrained Environments\footnoterule\thanks{$^{*}$ Corresponding author, email: ysguan@gdut.edu.cn. This work is partially supported by the National Natural Science Foundation of China (Grant No. 51975126), the Frontier and Key Technology Innovation Funds of Guangdong Province (Grant No. 2017B050506008, 2019B090915001) and the Guangdong Yangfan Program for Innovative and Entrepreneurial Teams (Grant No. 2017YT05G026).}
}

\author{\authorblockN{ Jiajie Ye, Yisheng Guan\textsuperscript{*}, Junfa Liu, Xinghong Huang and Hong Zhang}
\vspace{2mm}
\authorblockA{\textit{Biomimetic and Intelligent Robotics Lab (BIRL)}\\
\textit{School of Electro-mechanical Engineering}\\
\textit{Guangdong University of Technology}\\
\textit{Guangzhou, Guangdong Province, China}\\
}}%

\begin{document}

\maketitle
\thispagestyle{empty}
\pagestyle{empty}

\begin{abstract}
Person identification (P-ID) under real unconstrained noisy environments is a huge challenge. In multiple-feature learning with Deep Convolutional Neural Networks (DCNNs) or Machine Learning method for large-scale person identification in the wild, the key is to design an appropriate strategy for decision layer fusion or feature layer fusion which can enhance discriminative power. It is necessary to extract different types of valid features and establish a reasonable framework to fuse different types of information. In traditional methods, different persons are identified based on single modal features to identify, such as face feature, audio feature, and head feature. These traditional methods cannot realize a highly accurate level of person identification in real unconstrained environments. The study aims to propose a fusion module to fuse multi-modal features for person identification in real unconstrained environments. 

\end{abstract}


\section{INTRODUCTION}

    \begin{table*}[htbp]
    \centering
    \caption{Datasets for person identification. There are many dataset for face recognition and some for speaker recognition, but most of them only focus on one modality to complete the task of P-ID. Recently, some projects have opened source multi-modal datasets.}
    \label{dataset}
    \begin{tabular*}{18cm}{llllll}
    \hline
	Dataset & Task & Identities & Format & Clips/Face tracks & Images/Frames \\
	\hline
	LFW\cite{Huang2008Labeled} & Face Recog. & 5K & Image & - & 13K \\
	Megaface\cite{Miller2015MegaFace} & Face Recog. & 690K & Image & - & 1M \\
	MS-Celeb-1M\cite{Guo2016MS} & Face Recog. & 100K & Image & - & 10M \\
	YouTube Celebrities\cite{Kim2008Face} & Face Recog. & 47 & Video & 1,910 & - \\
	YouTube Faces\cite{Wolf2011Face} & Face Recog. & 1,595 & Video & 3,425 & 620K \\
	Buffy the Vampire Slayer\cite{Sivic2009} & Face Recog. & Around 19 & Video & 12K & 44K \\
    Big Bang Theory\cite{Bauml2013Semi} & Face $ \verb|&|$ Speaker Recog. & Around 8 & Video & 3,759 & - \\
	Sherlock\cite{Nagrani2018From} & Face $ \verb|&|$ Speaker Recog. & Around 33 & Video & 6,519 & - \\
	VoxCeleb\cite{Nagrani2017VoxCeleb} & Speaker Recog. & 6,112 & Video & 150K & - \\
	\hline
	iQIYI-VID-2019\cite{IEEEexample:articleetal} & Search & 10K & Video & 200K & 4M \\
	\hline
    \end{tabular*}
    \end{table*}
    
Most existing algorithms for person identification are based on quite constrained conditions since the partial datasets used to train are based on a constrained environment. Recently, there is an open-source video dataset based on the real unconstrained environment usually referred to as 'the wild' for person identification, iQIYI. Now, videos play a dominated role in the traditional media or new media since they contain multiple types of information, such as images, audio and text message. Consequently, video understanding has been explored for completing multiple tasks, especially P-ID. The task is mainly solved by face recognition, speaker recognition or any other biometric identification methods. The application scope of P-ID is wide and includes the authentication in high-security systems and forensic tests, and searching for persons in large corpora of video data. All such tasks require high multi-features learning performance under 'real world' conditions.

In video analysis, each topic addresses a single modal of information. As deep learning has been developing rapidly in recent years, all these video understanding methods have achieved great success. In the field of face recognition, based on the LFW benchmark \cite{Huang2008Labeled}, ArcFace \cite{Deng2018ArcFace} realized a precision of 99.83$ \verb|%|$, which had exceeded the human performance. The best results on Megaface \cite{Miller2015MegaFace} also reached 99.39$ \verb|%|$. In the field of speaker recognition, the Classification Error Rates of SincNet \cite{Ravanelli2018Speaker} based on the TIMIT dataset \cite{Garofolo1993DARPA} and LibriSpeech dataset \cite{Panayotov2015Librispeech} were only 0.85$ \verb|%|$ and 0.96$ \verb|%|$, respectively.

Everything seems right until these P-ID methods are applied in real unconstrained videos. Face recognition is sensitive to pose, blur, occlusion, etc. Moreover, in many video frames, faces are invisible, thus largely increasing the difficulty in face recognition. In person re-identification (Re-ID), the problem of changing clothes has not been considered yet. In the field of speaker recognition, one major challenge comes from the fact that the person to be recognized is not always speaking. Generally speaking, every single technique is inadequate to solve all the cases. Intuitively, the combination of all these sub-tasks together can fully utilize the rich contents of videos \cite{IEEEexample:articleetal}.

Some open source datasets are available to solve the P-ID task (Table \ref{dataset}). The available video datasets mainly utilize a single modality of feature, either face, audio, or body.

Based on different modalities of features, we used a fusion module to fuse multi-modal features for person identification in real unconstrained environments. We investigated different architectures and techniques for training deep CNNs with the features directly extracted from raw video files with little pre-processing and analyzed the performances of several state-of-the-art methods of single modal and multi-modal on iQIYI-VID dataset.

\section{Related Works}

\subsection{Face Recognition}

The task of face recognition can be divided into two sub-tasks, face verification and face identification. Face verification is a 1-to-1 matching problem of verifying whether the two given images belong to the same person. In 2007, the Labeled Faces in the Wild (LFW) dataset established for face verification by Huang et al. \cite{Huang2008Labeled} has become the most popular benchmark for face verification. Nevertheless, face recognition is a 1-to-k matching problem of recognizing whether a given image belongs to the image dataset containing k images. Based on LFW, many algorithms \cite{Schroff2015FaceNet,DBLP:journals/corr/SunLWT15,Yi2015Deeply,Taigman2014DeepFace} realized the recognition rate above 99$ \verb|%|$, which was better than the human performance \cite{Hu2015When}. The state-of-art method, ArcFace \cite{Deng2018ArcFace} achieved a face verification accuracy of 99.83$ \verb|%|$ based on LFW. In this study, ArcFace was adopted to recognize faces in videos.

For the purpose of enriching the contents of features, video datasets are a better choice in multi-modal P-ID. There are some video datasets, such as YouTube celebrity recognition dataset \cite{Kim2008Face} that includes videos of only 35 celebrities, and the YouTube Face Database (YFD) \cite{Wolf2011Face} that contains 3425 videos of 1595 persons. The biggest one is iQIYI-VID dataset that even contains some videos without visible faces.

\subsection{Speaker Identification}

The applications of speech processing technology are primarily classified as: speech recognition and speaker recognition. Speech recognition is to identify the spoken words, while speaker recognition is to identify speaker on the basis of his/her voice characteristics \cite{Campbell1997Speaker}. Speaker recognition is further dissected into two categories, speaker verification and speaker identification. Speaker verification is the process of validating the claim of identity by a speaker and consequently this type of decision is binary, i.e., true or false. In speaker identification, since there is no prior claim of an identity, the system classifies the input tested speech signals into one of the 'N' reference speakers. Speaker identification stated above is labeled as 'closed-set' speaker identification, which is different from 'open-set' identification, as in the case of open-set, the test speech signal may not belong to any of the 'N' reference speakers and N+1 decisions exist, thus leading to an additional result of the test signal not appertaining to any of the N reference speakers \cite{Doddington1985Speaker}.

Currently, speaker recognition still faces a dearth of freely available large-scale datasets in the wild. Some datasets, which were originally intended to be applied in speech recognition, such as TIMIT \cite{Garofolo1993DARPA} and LibriSpeech \cite{Panayotov2015Librispeech}, have been adopted in speaker recognition experiments. Many of these datasets were collected under controlled conditions and therefore improper for evaluating models under real conditions. To fill the gap, the Speaker in the Wild (SITW) dataset \cite{SITW} were generated from open multi-media resources. To the best of our knowledge, the largest and freely available speaker recognition datasets are VoxCeleb \cite{Nagrani2017VoxCeleb} and VbxCeleb2 \cite{Chung2018VoxCeleb2}.

\subsection{Head Feature Based Hairstyle Classification}

Despite audio recognition and face recognition can solve most of the problems of person identification, there are still some loopholes to increase the robustness of existing person identification algorithms. One of the limitations of most existing algorithms is the incapability to detect the presence of human beings under fully unconstrained conditions. Especially, if it is required to detect the presence of human beings from the back or over-the-shoulder views, without clear head-and-shoulder profiles, only human hair or accessories are available. Furthermore, hair contains the characteristics of textures and style and is one of the definite characteristics of human beings since it represents different cultures, historical periods, personal characteristic, gender and age. Hair detection in images is useful for face recognition, person identification, gender classification \cite{Li2012Gender}, and head detection in surveillance applications \cite{Zhang2009Head}.

\section{PROPOSED METHOD} 
The proposed method can be divided into two parts: features extraction and features fusion.
\subsection{Features extraction} 
\subsubsection{Face}
With the PyramidBox \cite{DBLP:journals/corr/abs-1803-07737}, one of the best face detectors, we could detect faces in the video frame. To improve the performance of the detector, the VGG16 backbone network of the PyramidBox is replaced with resnet50. In this way, the face recognition model ArcFace \cite{Deng2018ArcFace} can be used for feature extraction with VGG16.

According to our observations, satisfactory results cannot be obtained with only face features. Thus, two coefficients (detection score and quality score) are introduced to obtain the weighted average for the combined prediction. Firstly, the detection score is the highest score ranked based on the confidence of face detector. Some video clips may have more than one face with their correspondent bounding boxes. The bounding box with the biggest size because is selected since the features of only one face are required to predict the identification. Secondly, the quality score is the score simply ranked based on the L2- norm results of the faces output by the FC layer in the face recognition model of SphereFace \cite{Liu2017SphereFace}. Ranjan et al. observed that the L2-norm of the features learned with softmax loss was informative for measuring the face quality \cite{Ranjan2017L2}. In our experiments, the faces, which were regarded as low-quality faces, were mostly blurred faces, side faces, faces with partial error, and even invisible faces.

\subsubsection{Audio}
All audios from the video clips are first converted into single-channel16-bit streams at a sampling rate of 16 kHz for consistency. Spectrograms are then generated in a sliding window fashion under the hamming window width of 25 ms, a step of 10 ms and 512-point Fast Fourier Transform (FFT). Then, inspired by the previous report \cite{Nagrani2017VoxCeleb}, these spectrograms are used as the input to the CNN model of ResNet34 \cite{he2016deep}, which is trained as a classification model with the Voxceleb2 dataset \cite{Chung2018VoxCeleb2}. The 512D output from the last hidden layer is used as speaker embedding.

\subsubsection{Head}
Head features are a strong support to increase the robustness of the algorithm of P-ID in the wild and contain information from hair texture, hair style and accessories.

The head detector is YOLOv3 \cite{Redmon2018YOLOv3}. In some video clips, there are multiple persons mapped to multiple bounding boxes for heads, but only one major character is required. In order to acquire the major character, a major head is defined as a video frame, in which the biggest head is detected. After segmenting the head from a video frame, those head pictures will be resized and normalized as the input to be transferred to a head classifier based on the ArcFace model \cite{Deng2018ArcFace} and the previous report \cite{Svanera2016Figaro}.

\begin{figure}
\includegraphics[scale=0.4]{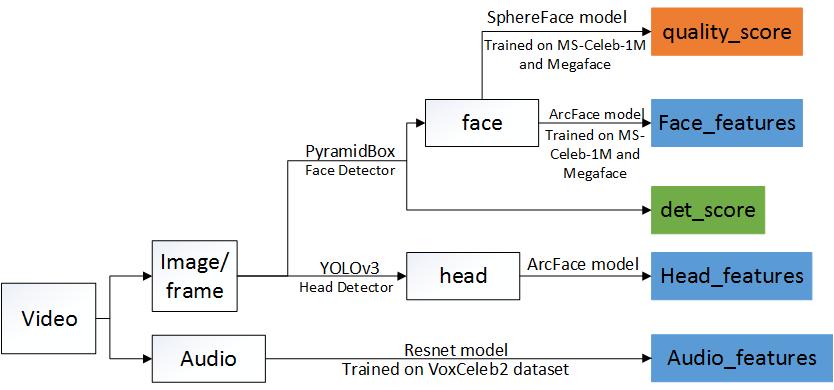}
\caption{Feature extraction. It is necessary to extract three representative and effective features with multiple detectors and models trained on different datasets.}
\label{extract_features}
\end{figure}

In general, the feature extraction procedure from face, audio and head are shown in Fig. \ref{extract_features}.

\subsection{Feature fusion}
\begin{figure*}
\centering
\includegraphics[scale=0.52]{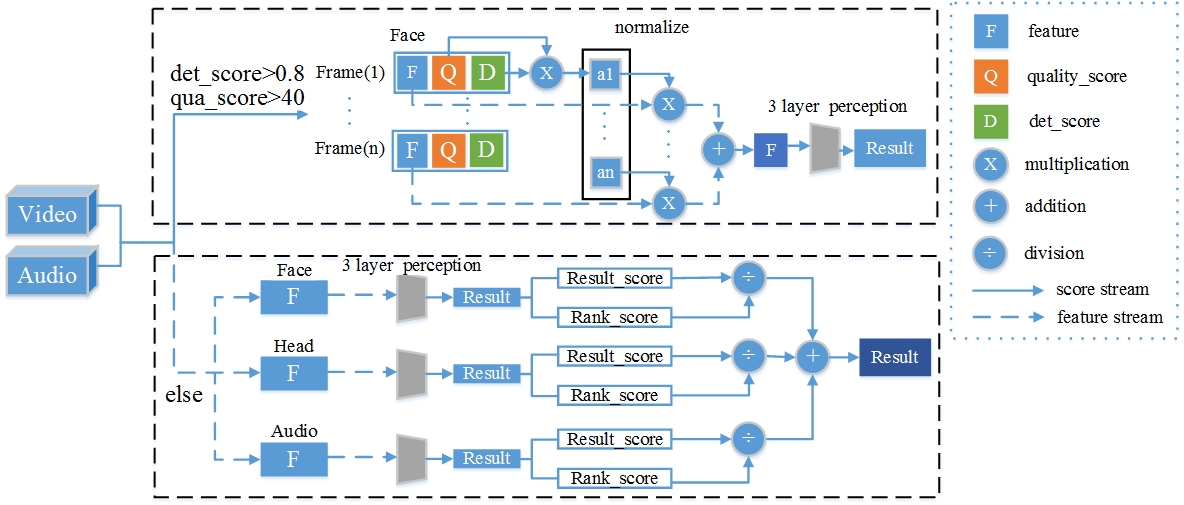}
\caption{Fusion strategy. According to the two coefficients of quality score and detection score, the videos with the high synthetic scores are input to the Part A module and the remaining videos are input to the Part B module.}
\label{fusion}
\end{figure*}

The fusion strategy of processing the input videos can be divided into two parts (Part A and Part B). Firstly, a single model of face is utilized to recognize the person identification based on the high-quality score and detection score, as shown in Part A in Fig. \ref{fusion}. Secondly, with the information from face, audio and head, the person identification is recognized in low-quality score and detection score, as shown in Part B of Fig. \ref{fusion}. The details of the fusion strategy are shown in Fig. \ref{fusion}.

\begin{itemize}

\item Part A represents the high-score partition and Part B represents the low-score partition.
\item According to each video's quality scores and detection scores, the videos with high scores of quality and detection are allocated to Part A for face recognition and other video clips are allocated to Part B.
\item In Part A, each video has an unknown number of frames, which contain face features, quality score and detection score. First, through multiplying quality score by detection score, the weight can be obtained, as shown in Eq. (1). Second, all the weights are normalized. Third, with the weight and their corresponding face features, the weighted average can be calculated. Fourth, as shown in Eq. (2), only one feature represents one video by integrating all the frames of one video based on the weighted average and those features are transferred to a classifier of 3-layer Multi-Layer Perception (MLP).
$$Normalize=
\begin{pmatrix}
a_1=qua\ score_1*det\ score_1 \\
a_2=qua\ score_2*det\ score_2 \\
\vdots \\
a_n=qua\ score_n*det\ score_n 
\end{pmatrix} \eqno(1)$$
$$F=\dfrac{\sum_{i=1}^n f_i*a_i}{\sum_{i=1}^n a_i} \eqno(2) $$
where $Normalize$ is normalizing inputs; $F$ is the feature representing a video; $n$ is the number of the video's frame; $a_1, \dots ,a_n $ is the result of multiplying quality score by detection score; $qua\ score$ is the quality score; $det\ score$ is the detection score; $f_i$ is the feature of video frame.
\item In Part B, single modal feature is used to recognize the P-ID through 3-layer MLP, and their results are fused to predict the result of low-score videos. The key of recognizing P-ID based on low-score videos is the fusion strategy. After the predicted results of three models are obtained, each predicted result has two parameters: result score and rank score. According to the predicted results, we choose the top 100 predicted results based on the ranked confidence. These 100 predicted results corresponded to 100 video’s ID. Consequently, each of the three models has 100 predicted results. As indicated in Eq. (3), according to the label, after the result score divided by rank score, the weighted score is acquired. However, a certain video ID is not necessarily contained in predicted results of all the three models at the same time. Thus, the weighted score obtained by fusing the predicted results involves three modals and sometime two modals. This case is similar to the real situation. For example, sometimes we only hear the sound, but we cannot see anyone, indicating the lack of visual features. Finally, as shown in Eq. (4), the final result is arranged according to the weighted score.
$$ Label\ i: \ W=\sum_{j=1}^m \dfrac{result\ score_j}{rank\ score_j} \eqno(3) $$
$$Result=
\begin{pmatrix}
label\ 1:\ W_{11},W_{12},\dots,W_{1k} \\
label\ 2:\ W_{21},W_{22},\dots,W_{2k} \\
\vdots \\
label\ N:\ W_{N1},W_{N2},\dots,W_{Nk} \\
\end{pmatrix} \eqno(4)$$
where $W$ is the weighted score of a video's ID; $result\ score$ is the confidence to match a label; $rank\ score$ is the sort of result score in top 100; $m$ is the number of the same video IDs in the same label from different modals; $N$ is the number of person ID; $k$ is that the top k results for each person ID.
\item After integrating the results of Part A and Part B, the final prediction result of the testing set is obtained.

\end{itemize}

\begin{figure*}
\centering
\includegraphics[scale=0.25]{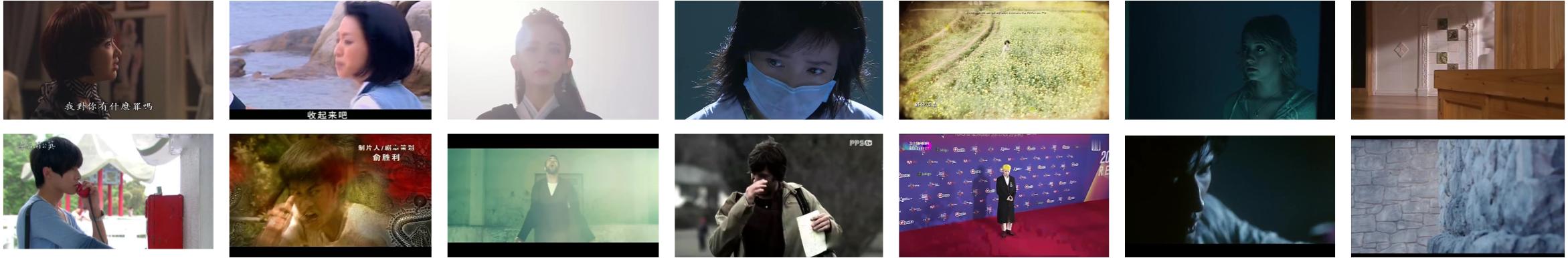}
\caption{Challenging cases for face recognition. From left to right: profile, blur, exposure, occlusion, small face, dark face, and invisible face.}
\label{dataset_sample}
\end{figure*}

\section{DATASET}
\begin{figure}
\centering
\subfigure[]{
\begin{minipage}[b]{0.5\textwidth}
\includegraphics[width=1\textwidth]{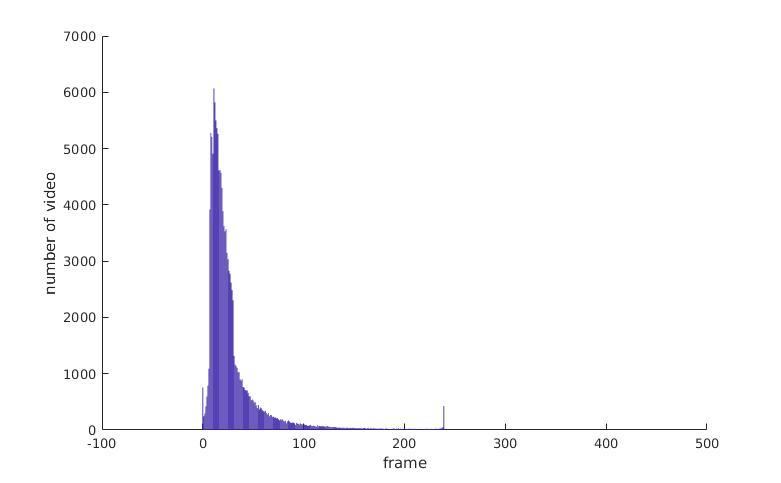}
\end{minipage}
}
\subfigure[]{
\begin{minipage}[b]{0.5\textwidth}
\includegraphics[width=1\textwidth]{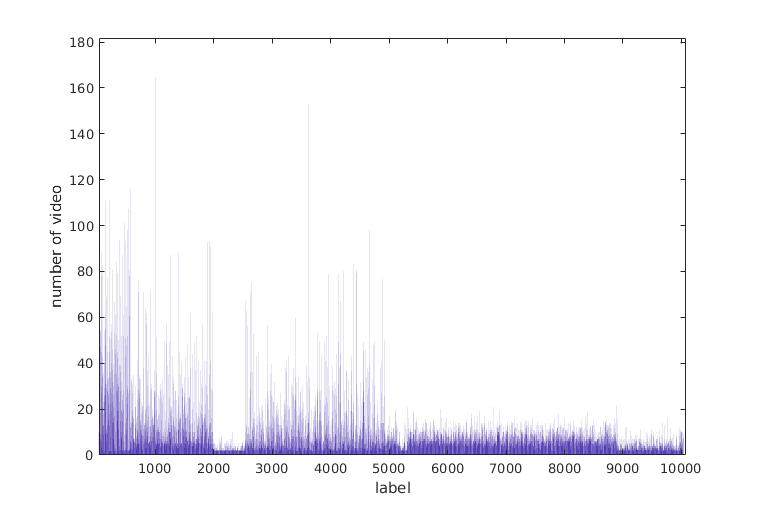}
\end{minipage}
}
 \caption{Distribution of the datasets. (a)The number of frame for a video. (b)The number of video for a label.} \label{distribution:dataset}
\end{figure}

In this study, the iQIYI-VID dataset is selected \cite{IEEEexample:articleetal} because it is a large-scale dataset that addresses the problem of multi-modal person identification. Especially, there are video clips without speakers or with only asides. The dataset contains 200 K video clips, which are divided into three parts: 40$ \verb|%|$ for training, 30$ \verb|%|$ for validation, and 30$ \verb|%|$ for testing. The dataset contains about 10, 000 identities. To mimic the real environment of video understanding, distracter videos with unknown person identities which are different from the major identities in the training set are inserted into the validation set and testing set. All the videos are manually labeled and can be used a good benchmark to evaluate person identification algorithms. The video clip duration is in the range of 1~30 seconds with an average of 4.72 second, and the distribution of the number of frames for video clips and the number of videos for labeling are shown in Fig. \ref{distribution:dataset}.

Besides, the video clip is a challenge case for face recognition since it includes profile, blur, exposure, occlusion, small face, dark face and invisible face. It is difficult to those faces with only face features. The samples are shown in Fig. \ref{dataset_sample}.

\section{EXPERIMENT}
\subsection{Experimental Preparation}
ArcFace \cite{Deng2018ArcFace} adopted raw features trained on MS-Celeb-1M \cite{Guo2016MS} and Megaface datasets \cite{Miller2015MegaFace}, and then input those raw features into a 3 layer Multi-Layer Perception (MLP). The pseudo-code of 3-layer MLP is shown in Table \ref{psedudocode}. Adam optimization algorithm with a learning rate of 0.0008 is adopted.

We combined the training set and validation set together and then divided it into 5 training sets, which were respectively used for training. Finally, 5 models respectively trained on 5 datasets were obtained.

In Part A, the high-score videos are divided into 4 groups according to the intervals of quality score, 40-200, 60-200, 80-200 and 100-200. Hence, the predicted results of Part A is a fusion of 20 models (each of the 5-fold dataset is divided into 4 groups).

In Part B, 3 types of features are used to predict the result under the low-score video frame. Note that, it is not necessary to divide the low-score videos because it is not a valid improvement operation. Hence, the predicted result of Part B is a fusion of 15 models (5 fold dataset for 3 types of feature models).

    \begin{table}[h]
    \caption{The pseudo-code of the 3 layer MLP}
    \label{psedudocode}
    \begin{tabular*}{8cm}{llll}
    \hline
	layer name & input & output & parameter  \\
	\hline
    FC1 & 512 & 1024 & activation='Relu' \\
    BN1 & - & - & batchnormalization,batch size=512 \\
    Drop1 & -& - & keep-prob=0.5 \\
    \hline
    FC2 & 1024 & 1024 & activation='Relu' \\
    BN2 & - & - & batchnormalization,batch size=512 \\
    Drop2 & -& - & keep-prob=0.5 \\
    \hline
    FC3 & 1024 & 10035 & activation='softmax' \\
	\hline
    \end{tabular*}
    \end{table}
    
\subsection{Evaluation Metrics}

Mean Average Precision (MAP) \cite{evalution2008} is used to evaluate the retrieval results: 
$$ MAP(Q)=\frac{1}{\left| Q \right|}\sum_{i=1}^{\left| Q \right|} \frac{1}{m_i} \sum_{j=1}^{n_i} Precision(R_{i,j}) \eqno(5)$$
where $Q$ is the number of person IDs; $m_i$ is the number of positive examples for the $i$-th ID; $n_i$ is the number of positive examples within the top k retrieval results for the $i$-th ID; $R_{i,j}$ is the set of ranked retrieval results from the top until you get j positive examples. In our implementation, only top 100 retrievals are kept for each person ID.

The top $K$ accuracy is not used in the evaluation since the dataset contains many video clips of unknown identities, which make the top $K$ accuracy invalid.

\subsection{Result}

\begin{table} [h] \small
\caption{Comparison with state-of-the-art methods of single modality and multi-modalities on iQIYI dataset. Note that, in this experiment, multiple datasets were used to train the model and this task was not accomplished just based on iQIYI dataset.}
\renewcommand\arraystretch{1.3}
\label{result}
\setlength{\tabcolsep}{2mm}{
\begin{tabular}{c|c|c}
\hline
 
\multicolumn{2}{c|}{\multirow{1}{*}{Method}} & 
\multicolumn{1}{c}{\multirow{1}{*}{Metric}} \\

\hline  
Modality & Model & MAP($ \verb|%|$) \\  
\hline  
Face & ArcFace-MLP & 83.54 \\  
Head & ArcFace-MLP & 59.42 \\ 
Audio & Resnet & 24.30 \\  
\hline  
Face+Head & Concatenate features and MLP &  84.27 \\
Face+Head+Audio & Concatenate features and MLP & 84.69 \\	
\hline   
the state-of-art\cite{IEEEexample:articleetal} & Attention module & 89.70 \\
\hline 
\textbf{Ours} & Fusion strategies & \textbf{92.17} \\  
\hline
\end{tabular}}

\end{table}

The models were evaluated with the metric of $MAP$ based on the testing set of iQIYI-VID dataset. We compared these models with the state-of-art methods in Table \ref{result}. Our method achieved a MAP of 92.17$ \verb|%|$, which was 2.47$ \verb|%|$ higher than that of the current state-of-art method.

Besides, the proposed method integrated multiple models which were trained on the dataset with different quality scores but the same distribution and realized the higher performance than the method, which directly fused 3 types of features and reached a precision of 84.69$ \verb|%|$ in the performance of P-ID.

\section{CONCLUSIONS}
In this study, we introduced new architectures and fusion strategies for the task of person identification and demonstrated the state-of-the-art performance based on the iQIYI-VID dataset. Through the comparison with the state-of-art models, experimental results showed that our method fusing multi-modal features outperformed the state-of-art models. Moreover, the provided method, a fusion module to fuse multi-modal features for P-ID in real unconstrained environments, adopts the decision layer fusion based on multiple prediction models, thus improve the accuracy and robustness of P-ID.






\bibliographystyle{IEEEtran}
\bibliography{IEEEabrv,mylib}


\end{document}